\documentclass[letterpaper, 10 pt, conference]{ieeeconf}  

\IEEEoverridecommandlockouts                              
\usepackage{graphics} 
\usepackage{epsfig} 
\usepackage{mathptmx} 
\usepackage{times} 
\usepackage{amsmath} 
\usepackage{amssymb}  
\usepackage{hyperref}

\title{\LARGE \bf
Segmentation of Liver Lesions with Reduced Complexity Deep Models
}

 \author{Ram Krishna Pandey$^{1}$, Ashwin Vasan$^{2}$ and A G Ramakrishnan$^{1}$
 \thanks{$^{1}$Ram krishna Pandey and A G Ramakrishnan are with Indian Institute of Science, Bangalore, India        {\tt\small Emails: \{ramp,agr\}@iisc.ac.in}}%
 \thanks{$^{2}$ Ashwin Vasan is currently working as project assistant in MILE lab at the Indian Institute of Science, Bangalore, India
         {\tt\small ashwinvasan94@gmail.com}}%
}

\begin{document}

\maketitle
\thispagestyle{empty}
\pagestyle{empty}

\begin{abstract}

We propose a computationally efficient architecture that learns to segment lesions from CT images of the liver. The proposed architecture uses bilinear interpolation with sub-pixel convolution at the last layer to upscale the course feature in bottle neck architecture. Since bilinear interpolation and sub-pixel convolution do not have any learnable parameter, our overall model is faster and occupies less memory footprint than the traditional U-net. We evaluate our proposed architecture on the highly competitive dataset of 2017 Liver Tumor Segmentation (LiTS) Challenge. Our method achieves competitive results while reducing the number of learnable parameters roughly by a factor of 13.8 compared to the original UNet model.  

\end{abstract}

\section{Introduction}
Medical diagnosis can be assisted by expert systems developed solely to increase the accuracy of such diagnoses. The development of image processing techniques along with the rapid development in areas like machine learning and computer vision help in creating such expert systems that almost nearly match the accuracy of the expert human eye. Being one of the forms of cancer in the modern society, liver cancer remains to be cured and eradicated. Diagnosis by Oncologists usually involves assessment of computed tomography (CT) scans. These scans are three dimensional and volumetric in nature and can contain up to a thousand different 2D slices. It takes a lot of human effort and time to manually segment each slice and the process is thus prone to errors with the segmentation varying between different oncologists. Thus, it is highly desirable to have algorithms that can automatically perform segmentation of liver lesions. This is a crucial area within medical diagnosis, where computer vision has made some progress~\cite{resultsBellver,xiaohan,mila1}.

The task of image segmentation faces more challenges when extended to the domain of medical images. Lesion detection in CT scans is influenced by a number of factors such as shape, area, noise, position and magnitude. Traditional methods for automatic segmentation have relied on thresholding, intensity clustering and region growing and have suffered from poor performance due to the challenging nature of medical images~\cite{xiaohan}. Hence, there is a clear demand for machine learning models that perform well for detecting and segmenting lesions.

Recently, fully convolutional neural networks (FCNs), including 2D and 3D FCNs, serve as the backbone in many volumetric image segmentation. Deep learning based segmentation architecture is generally composed of a down-sampling block followed by an up-sampling block. Down-sampling block is used to extract coarse features and up-sampling block takes these features to the same size and resolution as the input and, optionally, a post-processing module (e.g. conditional random fields) to refine the model predictions. Architectures which achieve state-of-the-art results often contain millions of parameters, which require a lot of compute power to process.

The recent deep learning revolution has changed the landscape of computer vision. The vision community has rapidly developed and improved deep neural networks, which now surpass human performance~\cite{andrejblog} in object classification~\cite{senets} in natural images. These algorithms and models have only been recently ported over to the domain of medical images and are often based on an encoder-decoder architecture~\cite{segnet}, which is an extension of fully convolutional networks~\cite{fcn}. Powerful baseline models for medical image segmentation often include UNets~\cite{unet}, an evolved FCN architecture, which has often been used to achieve state of the art performance across various problem statements involving CT scans and MRI images.

Our goal in this paper is to develop a fast and efficient framework for medical image segmentation that can be optimized to run on mobile phones as in~\cite{mobilenet}. As small computing devices become cheaper and better, they become more accessable to people all around the world. The shortage of doctors in third world countries can be counteracted with the ability to run models capable of performing medical diagnosis. Lightweight models such as ours can be easily used to assist doctors and empower scientific diagnosis.
\\

\section{Related Work}
Driven by powerful, deep learning architectures~\cite{resnet,resnext,inception,vgg}, tasks involving dense predictions such as semantic segmentation, surface normal prediction and instance segmentation have seen great improvement and progress over the last few years. Most of these models are driven by an encoder-decoder framework~\cite{segnet}. Along with the high-level semantic features that are required for accurate classification, pixel-level dense predictions also require low-level geometric features that preserve spatial information. 

Most of the work reported on dense predictions differ in the decoder block, where the aim is to upsample back to the original resolution. The original FCN architecture~\cite{fcn} used skip connections to concatenate feature maps from the encoder block to the only upsampling layer in the architecture. The work of Badrinarayanan et. al.~\cite{segnet} save the pooling indices in the encoder block and copy them to the corresponding upsampling layer in the decoder block. Some other methods like the UNet~\cite{unet} copy the entire feature maps from the encoder and concatenate them to the decoder. They have proved their better performance by becoming the defacto architecture behind several winning solutions within the Kaggle community for challenges involving medical images and satellite images~\cite{carvana,distsatellite}. 

UNet based architectures are popular for medical image based problem statements. For lesion segmentation, Chirst et. al.~\cite{cascadedfcn} used a two-step approach containing two UNet architectures to attain highly competitive scores on the 3DIRCAD dataset. Some authors like Drozdzal et. al.~\cite{fcresnet} replace convolutional blocks in UNet with a ResNet block to perform segmentation of anatomical organs. Certain others use a 3D variant of UNet~\cite{threeunet} to perform convolutions in three dimensional space. However, such operations are GPU intensive and take a lot of processing power without any substantial gain in performance. To our knowledge, all three top contenders in the liver tumor segmentation challenge (LiTS) use two-dimensional variants of UNets. Li et. al.~\cite{sota} use a hybrid approach, where they use a pre-trained 3D model and combine it with a 2D architecture. Vorontsov et. al.~\cite{mila1} use two parallel, fully convolutional networks and jointly train the two architectures by employing short-ranged skip connections to attain competitive results.

Depthwise separable convolutions were proposed first as part of the Xception architecture~\cite{xception} and later used as a key component in the architecture proposed by Howard et. al.~\cite{mobilenet}, where the model achieves competitive scores on the ImageNet challenge with the number of parameters less than one tenth of the state-of-the-art models. The reduced model footprint makes such architectures ideal for mobile vision applications. 

Our approach uses components inspired by different developments in deep learning. It is a simple and efficient architecture that gives good results with a single model and yet runs faster than traditional architectures and is also optimized for mobile devices.

\section{Problem Definition}
\begin{figure*}
\centering
\includegraphics[width=1.0\textwidth,height=0.30\textheight]{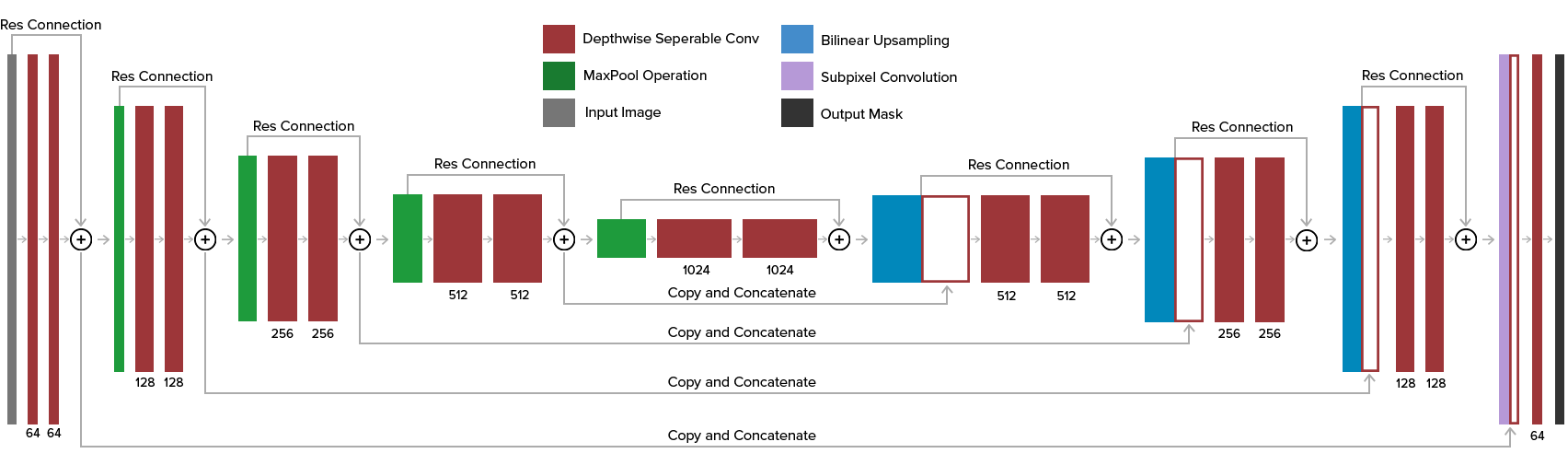}
\caption{Our proposed architecture consists of encoder-decoder modules, which include ResNet-blocks with residual skip connections. The model contains depthwise separable convolutions interspersed by downsampling layers in the encoder and upsampling layers in the decoder. Long-range skip connections facilitate the copy-and-concat mechanism that copy feature maps in the encoder and concatenate them to the corresponding feature map in the decoder.}
\label{fig: arch}
\end{figure*}

\subsection{Dataset}
The LiTS training dataset consists of 130 3D volumetric CT scans. These have been collected from different clinical sites and are contrast-enhanced for a better viewing experience. Different scanners produce CT scans that vary in terms of spatial resolution and field of view. For our convenience, we resize all the input slices to have a side-length of 256 pixels. The number of slices in each scan varies from around forty to over a thousand. 

\subsection{Preprocessing}
CT Windowing:\\
Windowing is frequently used in the evaluation of CT scans. We perform windowing of the CT Hounsfield units, also known as gray-level mapping, to a range from -100 to 200 HU. This provides a high soft-tissue contrast, which helps in the detection of tumor nodules, since it highlights certain structures.

Histogram Equalization:\\
Histogram equalization~\cite{histogramequal} is an image processing technique used to enhance contrast. The intensities are better distributed by spreading out their frequencies, thus spacing out values that are closely placed.

\section{Contribution}
Our method extends the UNet model~\cite{unet} by incorporating recent advances in deep learning architectures. Our main contributions are as follows.
\begin{itemize}
\item We have used short-range residual connection~\cite{resnet} for improved gradient flow, thus alleviating the vanishing gradient problem and resulting in a smooth and efficient training regime.

\item We advocate the use of depthwise separable convolution~\cite{xception}, which reduces the number of parameters used by a factor of over thirteen, when compared to the traditional UNet architecture, without compromising the performance.

\item We use a combination of bilinear interpolation~\cite{comparisoninterpolation} and subpixel convolution~\cite{subpixel} for upsampling back to the full resolution of the output mask in the decoder portion of the architecture. This totally avoids the use of slow deconvolution layer and hence reduces the number of learning parameters. This also helps in increasing the overall speed of training and inference time.
\end{itemize}

\section{Method}

\subsection{Architecture}
Our model consists of an encoder-decoder architecture and is based on UNet~\cite{unet} and employs long and short-range skip connections as in~\cite{mila2} and is end-to-end trainable. Figure~\ref{fig: arch} shows the architecture with all the layers and connections. All the convolution layers in the traditional UNet model are replaced by depthwise separable convolutions proposed in~\cite{xception}. These alternate type of convolutions achieve similar performance  as normal convolution operations with less number of parameters and thus are one of the key factors in reducing the model footprint.

Encoder:\\
The encoder receives a grayscale image as the input. It consists of four ResNet-like blocks~\cite{resnet}, where the input and output of the respective blocks are connected by a residual connection. The ResNet-like blocks are interspersed with a pooling layer, which performs the maxpooling operation to reduce the width and height by a factor of 2. Every ResNet block consists of two 2x2 depthwise separable convolution. The first convolution layer increases the depth dimension by doubling the input depth, while the second convolutions maintain the same depth as the previous layer. Any change in depth required by the residual connection is made by using a bottleneck layer proposed by He et. al. ~\cite{resnet}, which includes a $1 \times 1$ convolution with the necessary depth. As seen in Fig.~\ref{fig: arch}, the model starts with an initial depth of 64 and reaches a depth of 1024 in the innermost layer, where the model learns high-level semantic features.

Decoder:\\
The objective of the decoder is to obtain a high-resolution binary mask from the high-level semantic feature maps that the decoder receives as input through a series of layers and operations. This is achieved by 4 more ResNet-like blocks, where the convolutions are replaced by depthwise separable convolutions. These blocks function in the same way as they do at the encoder. These blocks are interspersed with an upsampling layer. The choice of an upsampling layer differs with the depth of the layer. The first 3 blocks in the decoder architecture are accompanied by a bilinear interpolation and the last upsampling is carried out by a subpixel convolutional layer~\cite{subpixel}. The latter layer simply rearranges the pixel values while increasing the spatial resolution and reducing the depth at the same time.

Skip Connections:\\
The architecture employs both long-range and short-range skip connections. As can be seen in Fig.~\ref{fig: arch}, the short-range skip connections are inspired by the ResNet architecture~\cite{resnet} and facilitate the flow of gradients. Thus aids the training process by directly addressing the ever persistent problem of vanishing gradients. The long-range skip connections are inspired by UNets~\cite{unet} and perform a copy-and-concatenate operation, where the feature maps are copied from the encoder to the corresponding decoder. These connections help in the accurate reconstruction of the mask, by copying the low-level feature maps from the encoder to the decoder and concatenate them to the corresponding upsampled layer. Convolving with these concatenated feature maps helps the model recover spatial information that was lost due to pooling operations within the encoder.

\subsection{Training and Implementation}
We trained the model using the Adam Optimizer with a learning rate of 0.001, $\beta_{1}=0.9$ and $\beta_{2}=0.999$. The model is evaluated using the dice score, also called the jaccard's coefficient and is defined as the ratio of the area of intersection to that of the union. We use the weighted cross entropy loss to train the model. Random rotations up to 180 degrees and elastic deformations with a zoom factor of 0.2 are applied as part of the data augmentation pipeline. 
\\ 
Batch normalization was applied before every ResNet block to address the internal covariate shift that affects deep learning models. A dropout of 0.05 was applied to facilitate training and to prevent the model from overfitting. We used k-fold validation and a split of 75:25 to train and validate our model.\\
We trained the model for 100,000 iterations with a batch size of 16. Training our model took approximately two days with GPU Titan X having 12 GB RAM, while training the basline UNet model took around four days. The final model weights are those, which yielded the best results on the validation set.

\begin{figure*}[!ht]
\centering
\includegraphics[width=.9\textwidth,height=0.25\textheight]{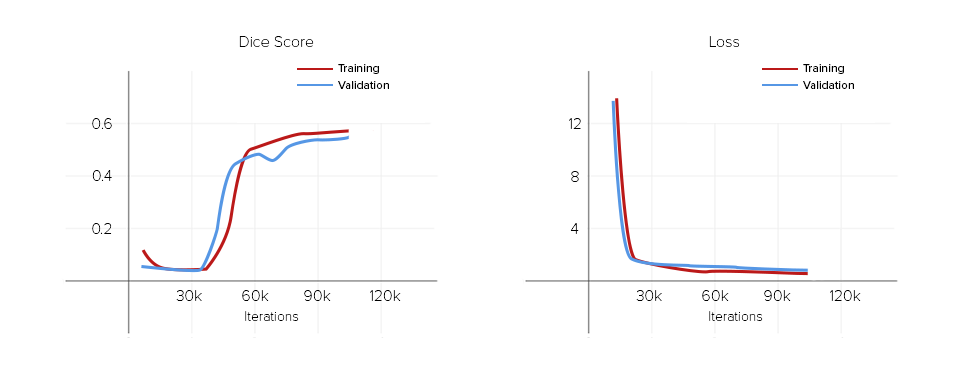}
\caption{ The training and validation curves along with the dice scores obtained after training the model for 100,000 iterations.The left and right figures are plots of dice score and the loss, respectively, as a function of the number of iterations.}
\label{fig: graph}
\end{figure*}

\section{Results and Discussion}
\begin{table}
\caption{Performance comparison of our model with submissions to LiTs challenge.}
 \resizebox{0.48\textwidth}{!}{
\begin{tabular}{ |l |c |  }
\hline
\hspace{3.3cm} \bf{Methods} & \bf{Dice} \\
 \hline
 Segmentation-only baseline~\cite{unet} & 0.41  \\  
 Segmentation-only 3-i/o+BP in liver~\cite{resultsBellver} & 0.54 \\
 Segmentation-only 3-i/o+BP in liver + Detector~\cite{resultsBellver} & 0.57\\
 Segmentation-only 3-i/o+BP in liver + Detector+3D-CRF~\cite{resultsBellver}& 0.59\\
 Vorontsov et. al~\cite{mila1} & 0.66 \\
 \hline
 \hspace{3.6cm}\bf{Ours} & 0.587 \\
 \hline

\end{tabular}
}
\end{table}
\vspace{0.45cm}

\begin{figure}[!ht]
\centering
\includegraphics[width=0.45\textwidth,height=0.23\textheight]{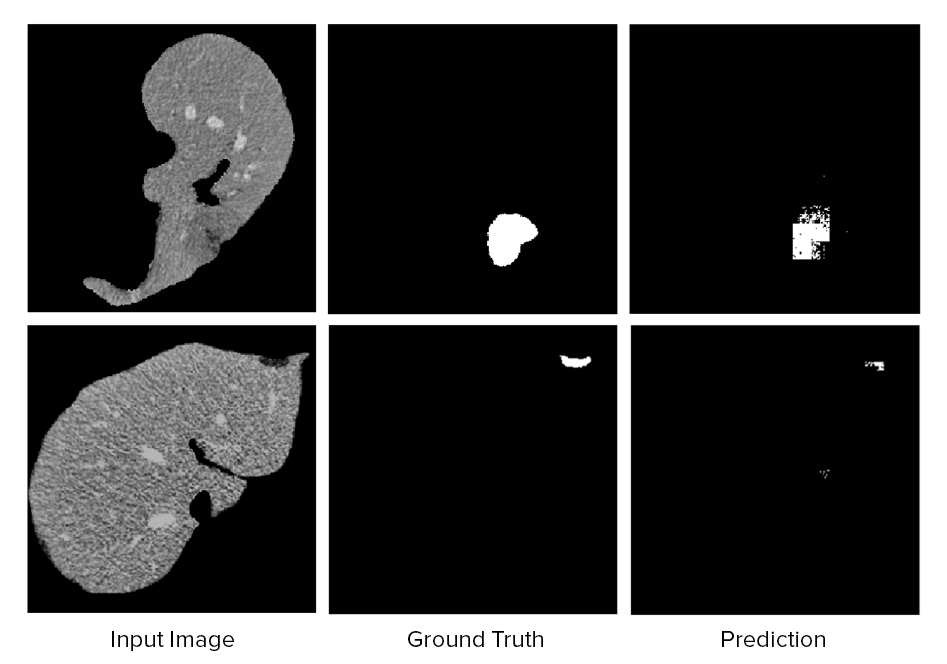}
\caption{Sample results obtained from the proposed architecture. Left figures: input images to our model; middle: ground truths; right figures: the results obtained from our segmentation model.}
\label{fig: result}
\end{figure}
Dice score can be calculated as:

\[
Dice\_Score = \frac{NTP}{TP+NFP} = \frac{|A\cap B|}{|A|+|B-A|}
\]
where, NTP or $A\cap B$ is the number of true positives and is like AND operation between the two sets (means the total number of pixels that have value 1 in both the ground truth (A) and the predicted image (B)), TP is true positives (total number of pixel having the intensity 1 in the ground truth) and NFP is the number of false positives (total number of pixels having intensity 0 in ground truth and 1 in the predicted image).
\\
The proposed method performs relatively well in the LiTS challenge, achieving competitive scores as shown in Table 1. Figure 3 shows an example segmentation result achieved by our model. Although other methods such as Vorontsov et. al.~\cite{mila1} achieve a higher dice score than our model, our method compares favorably in terms of the number of parameters used to achieve competitive scores. The baseline UNet model uses around 30 million learable parameters. Our model is able to achieve better performance with only 2 million learnable parameters, which leads to a 13.8x parameter reduction. In comparison, a MobileNet~\cite{mobilenet} implementation has around 4 million learnable parameters and is able to run efficiently on a mobile device. Thus we believe that our model can be implemented as a mobile application which can be used to perform medical diagnosis. Certain methods use pre-trained models, which may be hard to obtain when developing applications for similar real world problems. Our model does not depend on any pre-trained model or any form of post processing techniques such as 3D Conditional Random Fields~\cite{resultsBellver}, which are used by certain other submissions which achieve higher dice scores.

\section{Conclusion}
The major benefit of using an architecture such as ours is the reduced model footprint. Replacing deconvolution layers in the traditional UNet architectures with bilinear interpolations and subpixel convolutions greatly speeds up the training and inference time, while cutting down the number of learnable parameters. Using depthwise separable convolutions increases the overall non-linearity of the model, while reducing the number of parameters when compared to the normal convolutional operation. Employing long and short-ranged skip connections helps the training process and improves the model accuracy. The proposed architecture does not depend on any post-processing module and thus is end-to end trainable. The simplicity of the model architecture makes it a good base model for further research towards improving liver lesion segmentation.

\section{Future Work}
We wish to further explore other techniques for decoder upsampling. In particular, using other forms of interpolations may prove beneficial to recover finer details in the segmentation mask. Combinations of interpolation, subpixel and deconvolutional techniques can be further experimented with for exploring fast learnable upsampling. We believe that our model can easily be extended as a mobile application due to its reduced memory footprint and fast inference time. Deploying a deep learning model that gives good performance and is able to run as a native mobile application will truly empower diagnosis. We believe that our model can be easily extended and further optimized to perform faster operations.

\end{document}